\documentclass[11pt,a4paper]{article}
\usepackage[hyperref]{acl2018}
\usepackage{times}
\usepackage{latexsym}
\usepackage{amsfonts}
\usepackage{amsmath}
\usepackage{amssymb}
\usepackage{graphicx}
\usepackage{bbm}
\usepackage{lipsum}
\usepackage[shortlabels]{enumitem}
\usepackage{color}
\usepackage{algorithm}
\usepackage[noend]{algpseudocode}
\usepackage{url}
\usepackage[normalem]{ulem}
\usepackage[font={small}]{caption}
\usepackage{wrapfig}

\newcommand\sA{\ensuremath{\mathcal{A}}}

\newcommand\sZ{\ensuremath{\mathcal{Z}}}

\newcommand\FigStar[4]{\begin{figure*}[ht] \begin{center} \includegraphics[scale=#2]{#1} \end{center} \caption{\label{fig:#3} #4} \end{figure*}}

\newcommand\refsec[1]{Sec.~\ref{sec:#1}}

\newcommand\reffig[1]{Fig.~\ref{fig:#1}}

\newcommand\reftab[1]{Table~\ref{tab:#1}}

\aclfinalcopy 

\newcommand{\comment}[1]{}

\title{Weakly Supervised Semantic Parsing with Abstract Examples}

\author{Omer Goldman\thanks{$\quad$ Authors equally contributed to this work.}, \  Veronica Latcinnik\footnotemark[1], \ Udi Naveh\footnotemark[1], \  Amir Globerson, \ Jonathan Berant \\
Tel-Aviv University \\
\texttt{\{omergoldman@mail,veronical@mail,} \\ \texttt{ehudnave@mail,gamir@post,joberant@cs\}.tau.ac.il}
}

\date{}

\begin{document}
\maketitle
\begin{abstract}
Training semantic parsers from weak supervision (denotations) rather than strong supervision (programs) complicates training in two ways. First, a large \emph{search} space of potential programs needs to be explored at training time to find a correct program. Second, \emph{spurious} programs that accidentally lead to a correct denotation add noise to training.
In this work we propose that in closed worlds with clear semantic types, one can substantially alleviate these problems by utilizing an abstract representation, where tokens in both the language utterance and program are lifted to an abstract form. We show that these abstractions can be defined with a handful of lexical rules and that they result in  sharing between different examples that alleviates the difficulties in training.
To test our approach, we develop the first semantic parser for \textsc{CNLVR}, a challenging visual reasoning dataset, where the search space is large and overcoming spuriousness is critical, because denotations are either \textsc{True} or \textsc{False}, and thus random programs are likely to lead to a correct denotation. Our method substantially improves performance, and reaches 82.5\% accuracy, a 14.7\% absolute accuracy improvement compared to the best reported accuracy so far.

\end{abstract}

\section{Introduction} \label{sec:intro}

\begin{figure}[t]
\begin{center}
\includegraphics[scale=0.3]{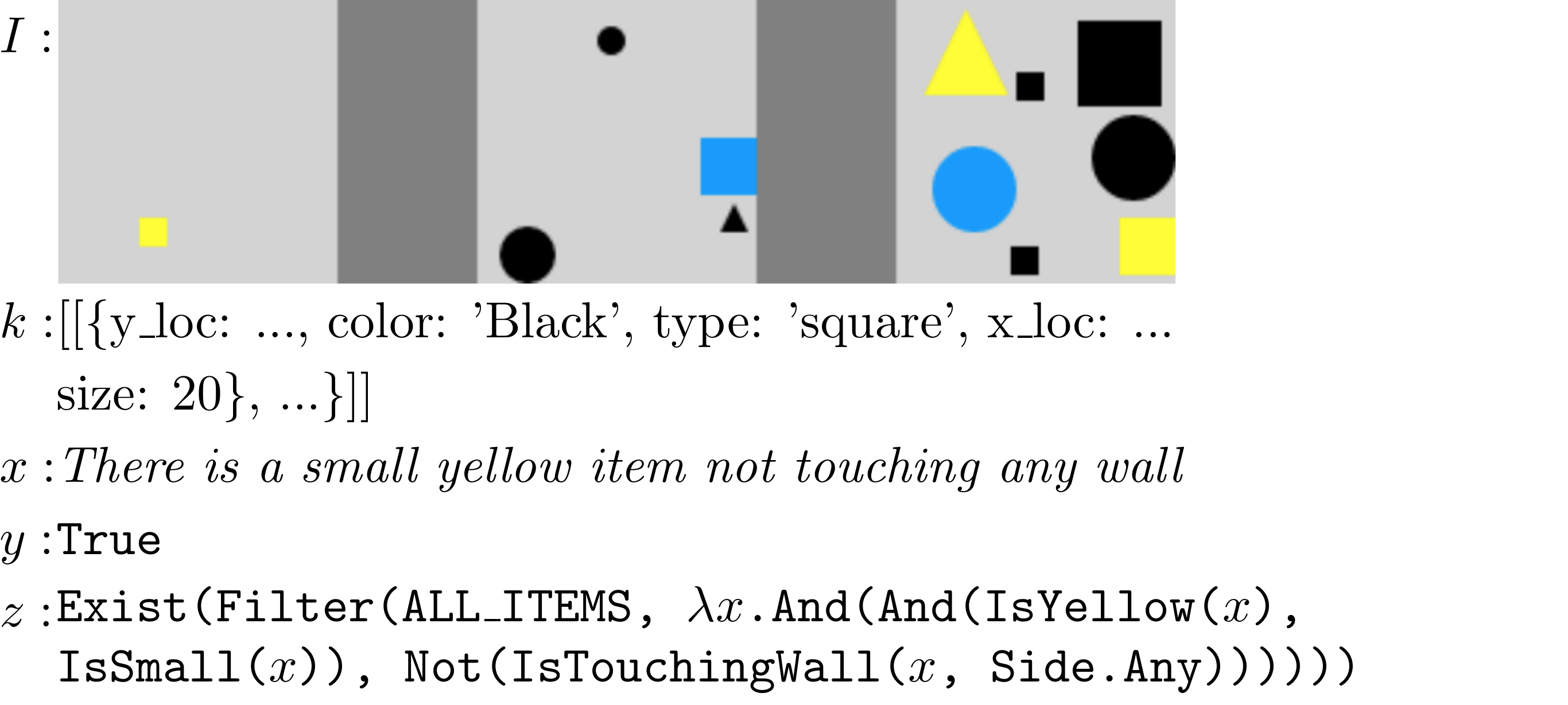}
\end{center}
\caption{Overview of our visual reasoning setup for the CNLVR dataset. Given an image rendered from a KB $k$ and an utterance $x$, our goal is to parse $x$ to a program $z$ that results in the correct denotation $y$. Our training data includes $(x,k,y)$ triplets.}
\label{fig:nlvr}
\end{figure}

The goal of semantic parsing is to map language utterances to executable programs. 
Early work on statistical learning of semantic parsers utilized supervised learning, where training examples included pairs of language utterances and programs
\cite{zelle96geoquery,kate05funql,zettlemoyer05ccg,zettlemoyer07relaxed}.
However, collecting such training examples at scale has quickly turned out to be difficult, because expert annotators who are familiar with formal languages are required. This has led to a  body of  work on weakly-supervised semantic parsing \cite{clarke10world,liang11dcs,krishnamurthy2012weakly,kwiatkowski2013scaling,berant2013freebase,cai2013large,artzi2013weakly}. In this setup,  training examples correspond to utterance-denotation pairs, where a \emph{denotation} is the result of executing a program against the environment (see \reffig{nlvr}). Naturally, collecting denotations is much easier, because it can be performed by non-experts.


Training semantic parsers from denotations rather than programs complicates training in two ways: (a) \emph{Search}: The algorithm must learn to search through the huge space of programs at training time, in order to find the correct program. This is a difficult search problem due to the combinatorial nature of the search space. (b) \emph{Spuriousness}: Incorrect programs can lead to correct denotations, and thus the learner can go astray based on these programs. Of the two mentioned problems, spuriousness has attracted relatively less attention \cite{pasupat2016inferring,guu2017bridging}.

Recently, the Cornell Natural Language for Visual Reasoning corpus (\textsc{CNLVR}) was released \cite{suhr2017nlvr}, and has 
presented an opportunity to better investigate the problem of spuriousness. In this task, an image with boxes that contains objects of various shapes, colors and sizes is shown. Each image is paired with a complex natural language statement, and the goal is to determine whether the statement is true or false (\reffig{nlvr}). 
The task comes in two flavors, where in one the input is the image (pixels), and in the other it is the knowledge-base (KB) from which the image was synthesized.
Given the KB, it is easy to view \textsc{CNLVR} as a semantic parsing problem: our goal is to translate language utterances into programs that will be executed against the KB to determine their correctness \cite{johnson2017inferring,hu2017learning}. Because there are only two return values, it is easy to generate programs that execute to the right denotation, and thus spuriousness is a major problem compared to previous datasets.

\begin{table*}[t]
\centering
\small
\resizebox{1.0\textwidth}{!}{
\begin{tabular}{l}
$x$: \emph{``There are exactly 3 yellow squares touching the wall.''} \\
$z$: \texttt{Equal(3, Count(Filter(ALL\_ITEMS, $\lambda x.$ And (And (IsYellow($x$), IsSquare($x$), IsTouchingWall($x$))))))} \\
\hline
$\bar{x}$: \emph{``There are C-QuantMod C-Num C-Color C-Shape touching the wall.''} \\
$\bar{z}$: \texttt{C-QuantMod(C-Num, Count(Filter(ALL\_ITEMS, $\lambda x.$ And (And (IsC-Color($x$), IsC-Shape($x$), IsTouchingWall($x$))))))} \\
\end{tabular}}
\caption{
An example for an utterance-program pair $(x,z)$ and its abstract counterpart $(\bar{x},\bar{z})$
}
\label{tab:abs-ex}
\end{table*}

In this paper, we present the first semantic parser for \textsc{CNLVR}. Semantic parsing can be coarsely divided into a \emph{lexical} task (i.e., mapping words and phrases to program constants), and a \emph{structural} task (i.e., mapping language composition to program composition operators). Our core insight is that in closed worlds with clear semantic types, like spatial and visual reasoning, we can manually construct a small lexicon that clusters language tokens and program constants, and create a partially abstract representation for utterances and programs (\reftab{abs-ex}) in which the lexical problem is substantially reduced.
This scenario is ubiquitous in many semantic parsing applications such as calendar, restaurant reservation systems, housing applications, etc: the formal language has a compact semantic schema and a well-defined typing system, and there are canonical ways to express many program constants.

We show that with abstract representations we can  share information across examples and better tackle  the search and spuriousness challenges. By pulling together different examples that share the same abstract representation, we can identify programs that obtain high reward across multiple examples, thus reducing the problem of spuriousness. 
This can also be done at search time,
by augmenting the search state with partial programs that have been shown to be useful in earlier iterations. Moreover, we can annotate a small number of abstract utterance-program pairs, and automatically generate training examples, that will be used to warm-start our model to an initialization point in which search is able to find correct programs.

We develop a formal language for visual reasoning, inspired by
\newcite{johnson2017inferring}, and train a semantic parser over that language
from weak supervision, showing that abstract examples substantially improve
parser accuracy. Our parser obtains an accuracy of 82.5\%, a 14.7\% absolute
accuracy improvement compared to state-of-the-art. All our code  is publicly available at \url{https://github.com/udiNaveh/nlvr_tau_nlp_final_proj}.

\section{Setup}

\paragraph{Problem Statement}
Given a training set of $N$ examples $\{(x_i, k_i, y_i)\}_{i=1}^N$, where $x_i$ is an utterance, $k_i$ is a KB describing objects in an image and $y_i \in \{\textsc{True}, \textsc{False}\}$ denotes whether the utterance is true or false in the KB,
our goal is to learn a semantic parser that maps a new utterance $x$ to a program $z$ such that when $z$ is executed against the corresponding KB $k$, it yields the correct denotation $y$ (see \reffig{nlvr}).

\paragraph{Programming language}
\begin{table*}[t]
\centering
\small
\resizebox{1.0\textwidth}{!}{
\begin{tabular}{l}
$x$: \emph{``There is a small yellow item not touching any wall.''} \\
$z$: \texttt{Exist(Filter(ALL\_ITEMS, $\lambda x$.And(And(IsYellow($x$), IsSmall($x$)), Not(IsTouchingWall($x$, Side.Any)))))} \\
\hline
$x$: \emph{``One tower has a yellow base.''} \\
$z$: \texttt{GreaterEqual(1, Count(Filter(ALL\_ITEMS, $\lambda x$.And(IsYellow($x$), IsBottom($x$)))))}
\end{tabular}}
\caption{Examples for utterance-program pairs. Commas and parenthesis provided for readability only.
}
\label{tab:programs}
\end{table*}

The original KBs in \textsc{CNLVR} describe an image as a set of objects, where each object has a color, shape, size and location in absolute coordinates. 
We define a programming language over the KB that is more amenable to spatial reasoning, inspired by work on the \textsc{CLEVR} dataset \cite{johnson2017inferring}.
This programming language provides access to functions that allow us to check the size, shape, and color of an object, to check whether it is touching a wall, to obtain sets of items that are above and below a certain set of items, etc.\footnote{We leave the problem of learning the programming language functions from the original KB for future work.} More formally, a program is a sequence of tokens describing a possibly recursive sequence of function applications in prefix notation. Each token is either a function with fixed arity (all functions have either one or two arguments), a constant, a variable or a $\lambda$ term used to define Boolean functions. Functions, constants and variables have one of the following atomic types: \texttt{Int}, \texttt{Bool}, \texttt{Item}, \texttt{Size}, \texttt{Shape}, \texttt{Color}, \texttt{Side} (sides of a box in the image); or a composite type \texttt{Set(?)}, and \texttt{Func(?,?)}. Valid programs have a return type \texttt{Bool}. Tables \ref{tab:abs-ex} and \ref{tab:programs} provide examples for utterances and their correct programs.
The supplementary material provides a full description of all program tokens, their arguments and return types.

Unlike \textsc{CLEVR}, \textsc{CNLVR} requires substantial set-theoretic reasoning (utterances refer to various aspects of sets of items in one of the three boxes in the image), which required extending the language described by \newcite{johnson2017inferring} to include set operators and lambda abstraction.
We manually sampled 100 training examples from the training data and estimate that roughly 95\% of the utterances in the training data can be expressed with this programming language.

\section{Model} \label{sec:model}
We base our model on the semantic parser of \citet{guu2017bridging}. In their work, they used an encoder-decoder architecture \cite{sutskever2014sequence} to define a distribution $p_\theta(z \mid x)$. The utterance $x$ is encoded using a bi-directional LSTM \cite{hochreiter1997lstm} that creates a contextualized representation $h_i$ for every utterance token $x_i$, and the decoder is a feed-forward network combined with an attention mechanism over the encoder outputs \cite{bahdanau2015neural}. The feed-forward decoder takes as input the last $K$ tokens that were decoded. 

More formally the probability of a program is the product of the probability of its tokens given the history: $p_\theta(z \mid x) = \prod_t p_\theta(z_t \mid x, z_{1:t-1})$, and the probability of a decoded token is computed as follows. First, a Bi-LSTM encoder converts the input sequence of utterance embeddings into a sequence
of forward and backward states $h^{\{F,B\}}_1,\ldots,h^{\{F,B\}}_{|x|}$. The utterance representation $\hat{x}$ is $\hat{x} = [h_{|x|}^F ; h_1^B]$.
\comment{
\begin{align*}
h_i^F &= \text{LSTM}(h_{i-1}^F, \phi_x(x_i)), \\
h_i^B &= \text{LSTM}(h_{i+1}^B, \phi_x(x_i)), \\
h_i &= [h_i^F; h_i^B] \ , \ \hat{x} = [h_{|x|}^F ; h_1^B], \\
\end{align*}
where $\phi_x(\cdot)$ provides utterance embeddings and '$;$' corresponds to concatenation. 
}
Then decoding produces the program token-by-token:
\begin{align*}
&q_t = \text{ReLU}(W_q[\hat{x}; \hat{v}; z_{t-K-1: t-1}]), \\
&\alpha_{t,i} \propto \exp(q_t^\top W_\alpha h_i) \ , \ c_t = \sum_i \alpha_{t,i}  h_i, \\
&p_\theta (z_t \mid x, z_{1:t-1}) \propto \exp(\phi_{z_t}^\top W_s[q_t; c_t]),
\end{align*}
where $\phi_z$ is an embedding for program token $z$, $\hat{v}$ is a bag-of-words vector for the tokens in $x$, $z_{i:j} = (z_i, \dots, z_j)$ is a history vector of size K, the matrices $W_q, W_\alpha, W_s$ are learned parameters (along with the LSTM parameters and embedding matrices), 
and '$;$' denotes concatenation.

\paragraph{Search:}
Searching through the large space of programs is a fundamental challenge in semantic parsing. To combat this challenge we apply several techniques. First, we use beam search at decoding time and when training from weak supervision (see \refsec{training}), similar to prior work \cite{liang2017nsm,guu2017bridging}. At each decoding step we maintain a beam $B$ of program prefixes of length $n$, expand them exhaustively to programs of length $n+1$ and keep the top-$|B|$ program prefixes with highest model probability.

\FigStar{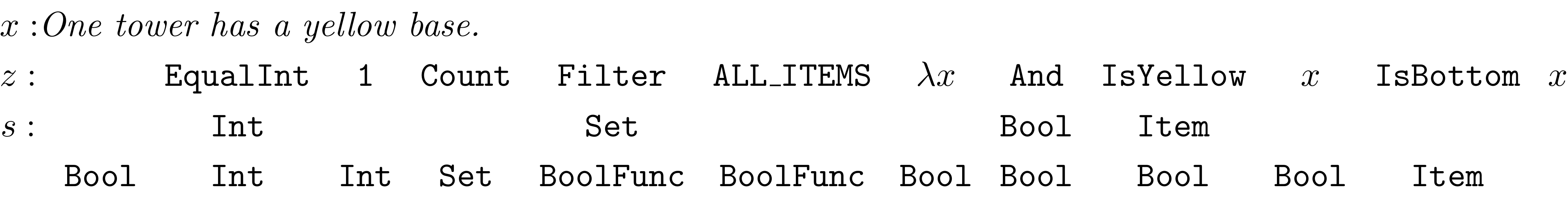}{0.35}{stack}{An example for the state of the type stack $s$ while decoding a program $z$ for an utterance $x$.}

Second, we  utilize the semantic typing system to only construct programs that are syntactically valid, and substantially prune the program search space (similar to type constraints in   \newcite{krishnamurthy2017neural,xiao2016sequence,liang2017nsm}). We maintain a stack that keeps track of the expected semantic type at each decoding step. The stack is initialized with the type \texttt{Bool}. Then, at each decoding step, only tokens that return the semantic type at the top of the stack are allowed, the stack is popped, and if the decoded token is a function, the semantic types of its arguments are pushed to the stack. This dramatically reduces the search space and guarantees that only syntactically valid programs will be produced. \reffig{stack} illustrates the state of the stack when decoding a program for an input utterance.


Given the constrains on valid programs, our model $p'_\theta(z \mid x)$ is defined as:
\begin{align*}
\prod_t \frac{p_\theta(z_t \mid x, z_{1:t-1}) \cdot \mathbbm{1}(z_t \mid z_{1:t-1})}{\sum_{z'} p_\theta(z' \mid x, z_{1:t-1}) \cdot \mathbbm{1}(z' \mid z_{1:t-1})},
\end{align*}
where $\mathbbm{1}(z_t \mid z_{1:t-1})$ indicates whether a certain program token is valid given the program prefix.

\paragraph{Discriminative re-ranking:}
The above model is a locally-normalized model that provides a distribution for every decoded token, and thus might suffer from the label bias problem \cite{andor2016globally,lafferty01crf}.
Thus, we add a globally-normalized re-ranker $p_\psi(z\mid x)$ that scores all $|B|$ programs in the final beam produced by $p'_\theta(z \mid x)$. Our globally-normalized model is:
$$p^g_\psi(z \mid x) \propto \exp(s_\psi(x,z)),$$
and is normalized over all programs in the beam. 
The scoring function $s_\psi(x,z)$ is a neural network with identical architecture to the locally-normalized model, except that (a) it feeds the decoder with the candidate program $z$ and does not generate it. (b) the last hidden state is inserted to a feed-forward network whose output is $s_\psi(x,z)$. 
Our final ranking score is $p'_\theta(z|x) p^g_\psi(z \mid x)$.


\section{Training}
\label{sec:training}

We now describe our basic method for training from weak supervision, which we extend upon in \refsec{learning} using abstract examples.
To use weak supervision, we treat the program $z$ as a latent variable that is approximately marginalized. To describe the objective, define $R(z,k,y)\in\{0,1\}$ to be one if executing program $z$ on KB $k$ results in denotation $y$, and zero otherwise. The objective is then to maximize $p(y \mid x)$ given by:
\begin{align*}
\sum_{z \in \sZ} p'_\theta(z \mid x)   p(y \mid z, k)
&= \sum_{z \in \sZ} p'_\theta(z \mid x)   R(z, k, y) \\ &\approx \sum_{z \in B} p'_\theta(z \mid x)  R(z, k, y)
\end{align*}
where $\sZ$ is the space of all programs and $B \subset \sZ$ are the
programs found by beam search.

In most semantic parsers there will be relatively few $z$ that generate the correct denotation $y$. However, in \textsc{CNLVR}, $y$ is binary, and so spuriousness is a central problem. To alleviate it, we utilize a  property of \textsc{CNLVR}: the same utterance appears 4 times with 4 different images.\footnote{
We used the KBs in \textsc{CNLVR}, for which there are 4 KBs per utterance. When working over pixels there are 24 images per utterance, as 6 images were generated from each KB.} If a program is spurious it is likely that it will yield the wrong denotation in one of those 4 images. 

Thus, we can re-define each training example to be  $(x, \{(k_j, y_j)\}_{j=1}^4)$, where each utterance $x$ is paired with 4 different KBs and the denotations of the utterance with respect to these KBs. Then, we maximize $p(\{y_j\}_{j=1}^4 \mid x, \comment{\{k_j\}_{j=1}^4})$ by maximizing the objective above, except that $R(z, \{k_j, y_j\}_{j=1}^4)=1$ iff the denotation of $z$ is correct for \textbf{all} four KBs. This dramatically reduces the problem of spuriousness, as the chance of randomly obtaining a correct denotation goes down from $\frac{1}{2}$ to $\frac{1}{16}$. This is reminiscent of \newcite{pasupat2016inferring}, where random permutations of Wikipedia tables were shown to crowdsourcing workers to eliminate spurious programs.

We train the discriminative ranker analogously by maximizing the probability of programs with correct denotation $\sum_{z \in B} p^g_\psi(z \mid x) R(z, k, y)$.

This basic training method fails for \textsc{CNLVR} (see \refsec{experiments}), due to the difficulties of search and spuriousness. Thus, we turn to learning from abstract examples, which substantially reduce these problems.


\section{Learning from Abstract Examples} \label{sec:learning}

\begin{table}[t]
\begin{center}
\scriptsize{
\begin{tabular}{l|l|l|l}
 \textbf{Utterance} & \textbf{Program} & \textbf{Cluster} & \textbf{\#} \\ 
 \hline \hline
 \emph{``yellow''} & \texttt{IsYellow} & \texttt{C-Color} & 3 \\
 \emph{``big''} & \texttt{IsBig} & \texttt{C-Size} & 3 \\
 \emph{``square''} & \texttt{IsSquare} & \texttt{C-Shape} & 4 \\
 \emph{``3''} & \texttt{3} & \texttt{C-Num} & 2 \\
 \emph{``exactly''} & \texttt{EqualInt} & \texttt{C-QuantMod} & 5\\
  \emph{``top''} & \texttt{Side.Top} & \texttt{C-Location} & 2 \\
    \emph{``above''} & \texttt{GetAbove} & \texttt{C-SpaceRel} & 6 \\
\hline
 &  & \textbf{Total:} & \textbf{25} \\
\end{tabular}}
\end{center}
\caption{Example mappings from utterance tokens to program tokens for the seven clusters used in the abstract representation. The rightmost column counts the number of mapping in each cluster, resulting in a total of 25 mappings.}
\label{tab:abstraction}
\end{table}


The main premise of this work is that in closed, well-typed domains such as visual reasoning, the main challenge is handling language compositionality, since questions may have a complex and nested structure. Conversely, the problem of mapping lexical items to functions and constants in the programming language 
can be substantially alleviated by taking advantage of the compact KB schema and typing system, and utilizing a small lexicon that maps prevalent lexical items into typed program constants.
Thus, if we abstract away from the actual utterance into a partially abstract representation, we can combat the search and spuriousness challenges as we can generalize better across examples in small datasets.

Consider the utterances:
\begin{enumerate}[nosep]

\item \emph{``There are exactly 3 yellow squares touching the wall."}
\item \emph{``There are at least 2 blue circles touching the wall."}

\end{enumerate}
While the surface forms of these utterances are different, at an abstract level they are similar and it would be useful to leverage this similarity.

We therefore define an abstract representation for utterances and logical forms that is suitable for spatial reasoning.
We define seven abstract clusters (see Table \ref{tab:abstraction}) that correspond to the main semantic types in our domain. Then, we associate each cluster with a small lexicon that contains language-program token pairs associated with this cluster. These mappings represent the canonical ways in which program constants are expressed in natural language.
\reftab{abstraction} shows the seven clusters we use, with an example for an utterance-program token pair from the cluster, and the number of mappings in each cluster. In total, 25 mappings are used to define abstract representations.

As we show next, abstract examples can be used to improve the process of training semantic parsers. Specifically, in sections \ref{sec:abs}-\ref{sec:caching}, we use abstract examples in several ways, from generating new training data to improving search accuracy. The combined effect of these approaches is quite dramatic, as our evaluation demonstrates.   

\subsection{High Coverage via Abstract Examples\label{sec:abs}}
We begin by demonstrating that abstraction leads to rather effective coverage of the types
of questions asked in a dataset. Namely, that many questions in the data correspond to a small set of abstract examples. We created abstract representations for all 3,163 utterances in the training examples by mapping utterance tokens to their cluster label, and then counted how many distinct abstract utterances exist. We found that as few as 200 abstract utterances cover roughly half of the training examples in the original training set. 

The above suggests that knowing how to answer a small set of abstract questions may already yield a reasonable baseline. To test this baseline, we constructured a ``rule-based'' parser as follows. We manually annotated 106 abstract utterances with their corresponding abstract program (including alignment between abstract tokens in the utterance and program). For example, \reftab{abs-ex} shows the abstract utterance and program for the utterance \emph{``There are exactly 3 yellow squares touching the wall"}. Note that the utterance \emph{``There are at least 2 blue circles touching the wall"} will be mapped to the same abstract utterance and program.

Given this set of manual annotations, our rule-based semantic parser operates as follows: Given an utterance $x$, create its abstract representation $\bar{x}$. If it exactly matches one of the manually annotated utterances, map it to its corresponding abstract program $\bar{z}$. Replace the abstract program tokens with 
real program tokens based on the alignment with the utterance tokens, and obtain a final program $z$. If $\bar{x}$ does not match return $\textsc{True}$, the majority label. The rule-based parser will fail for examples not covered by the manual annotation. However, it already provides a reasonable baseline (see Table \ref{tab:main_res}). As shown next, manual annotations can also be used for generating new training data.

\subsection{Data Augmentation \label{sec:augmented}}
While the rule-based semantic parser has high precision and gauges the amount of structural variance in the data, it cannot generalize beyond  observed examples. However, 
we can automatically generate non-abstract utterance-program pairs from the manually annotated abstract pairs and train a semantic parser with strong supervision that can potentially generalize better.
E.g., consider the utterance \emph{``There are exactly 3 yellow squares touching the wall"}, whose abstract representation is given in \reftab{abs-ex}. It is clear that we can use this abstract pair to generate 
a program for a new utterance \emph{``There are exactly 3 blue squares touching the wall"}. This program will be identical to the program of the first utterance, with \texttt{IsBlue} replacing \texttt{IsYellow}.

More generally, we can sample any abstract example and instantiate the abstract clusters that appear in it by sampling pairs of utterance-program tokens for each abstract cluster.
Formally, this is equivalent to a synchronous context-free grammar \cite{chiang2005hierarchical} that has a rule for generating each manually-annotated abstract utterance-program pair, and rules for synchronously generating utterance and program tokens from the seven clusters.

We generated 6,158 $(x,z)$ examples using this method and trained a standard sequence to sequence parser by maximizing $\log p'_\theta(z | x)$ in the model above. Although these are generated from a small set of 106 abstract utterances, they can be used to learn a model with higher coverage and accuracy compared to the rule-based parser, as our evaluation demonstrates.\footnote{Training a  parser directly over the 106 abstract examples results in poor performance due to the small number of examples.}

The resulting parser can be used as a standalone semantic parser. However, 
it can also be used as an initialization point for the weakly-supervised semantic parser. As we observe in  \refsec{experiments}, this results in further improvement in accuracy.

\subsection{Caching Abstract Examples \label{sec:caching}}

\begin{algorithm}[t] 
\caption{Decoding with an Abstract Cache}
\label{alg:cache} 
\begin{algorithmic}[1]
{\footnotesize

\Procedure {Decode}{$x,y,C,D$}
\State // C is a map where the key is an abstract utterance and the value
is a pair $(Z,\hat{R})$ of a list of abstract programs $Z$ and their average
rewards $\hat{R}$. $D$ is an integer.
\State $\bar{x} \leftarrow$ Abstract utterance of x
\State $\sA \leftarrow$ $D$ programs in $C[\bar{x}]$ with top reward values
\State $B_1 \leftarrow$ compute beam of programs of length $1$
\For{$t = 2 \dots T$} // Decode with cache
\State $B_t \leftarrow$ construct beam from $B_{t-1}$
\State $\sA_{t} = \text{truncate}(\sA, t)$
\State $B_t.\text{add}(\text{de-abstract}(\sA_{t}))$ \label{line:prefix}
\EndFor
\For{$z \in B_T$} //Update cache \label{line:cacheupdate}
\State Update rewards in $C[\bar{x}]$ using $(\bar{z}, R(z,y))$
\EndFor
\State return $B_T \cup
\text{de-abstract}(\sA)$.\label{line:full}
\EndProcedure
}
\end{algorithmic}
\end{algorithm}

\FigStar{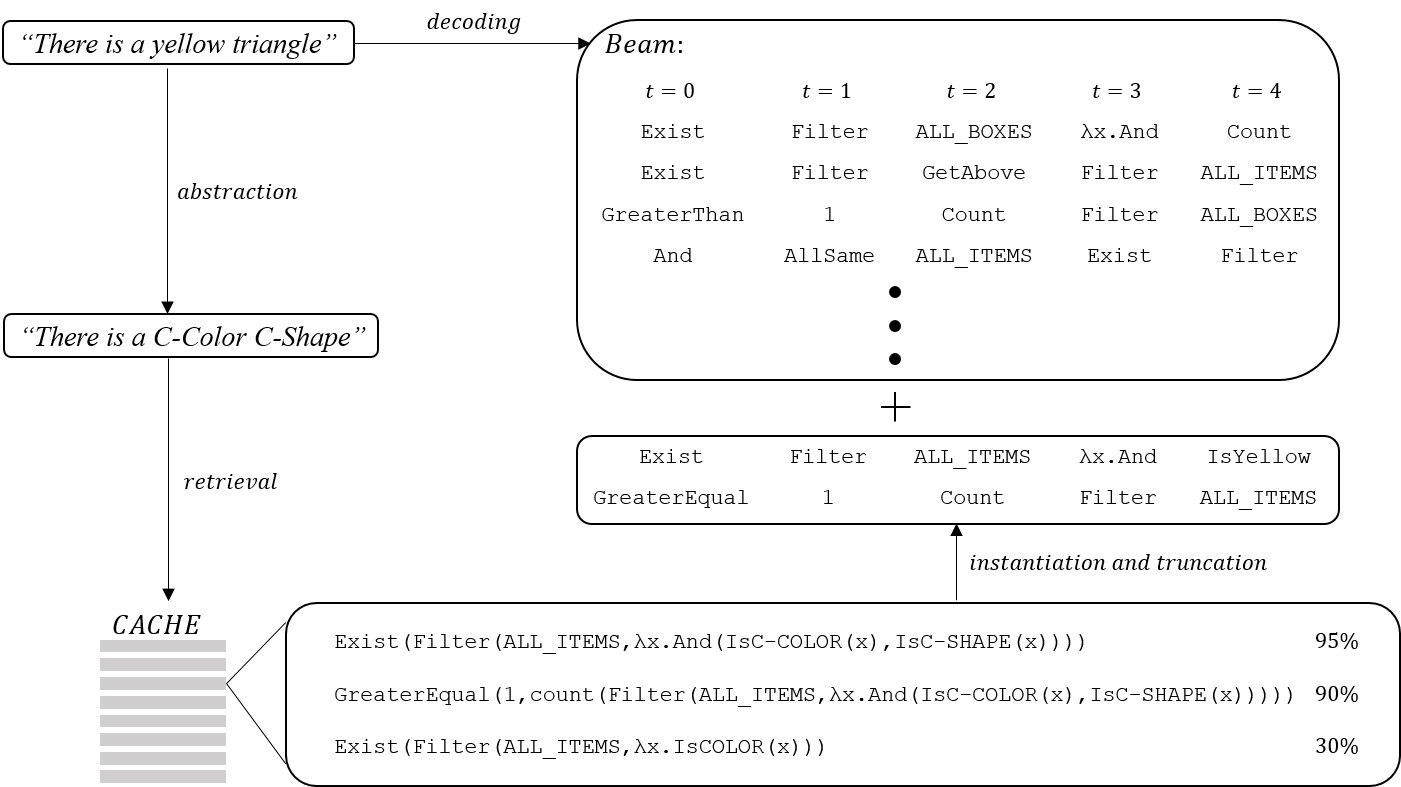}{0.55}{cache}{ A visualization of the caching mechanism. 
At each decoding step, prefixes of 
high-reward abstract programs are added to the beam from the cache.}

We now describe a caching mechanism that 
uses abstract examples to combat search and spuriousness when training from weak supervision.
As shown in \refsec{abs}, many utterances are identical at the abstract level. Thus, a natural idea is to keep track at training time of abstract utterance-program pairs that resulted in a correct denotation, and use this information to direct the search procedure.


Concretely, we construct a cache $C$ that maps abstract utterances to all abstract programs that were decoded by the model, and tracks the average reward obtained for those programs. For every utterance $x$, after obtaining the final beam of programs, we add to the cache all abstract utterance-program pairs $(\bar{x}, \bar{z})$, and update their average reward (Alg.~\ref{alg:cache}, line \ref{line:cacheupdate}). 
 To construct an abstract example $(\bar{x}, \bar{z})$ from an utterance-program pair $(x,z)$ in the beam, we perform the following procedure. First, we create $\bar{x}$ by replacing utterance tokens with their cluster label, as in the rule-based semantic parser. Then, we go over every program token in $z$, and replace it with an abstract cluster if the utterance contains a token that is mapped to this program token according to the mappings from \reftab{abstraction}. This also provides an alignment from abstract program tokens to abstract utterance tokens that is necessary when utilizing the cache. 

We propose two variants for taking advantage of the cache $C$. Both are shown in Algorithm \ref{alg:cache}.
\newline
{\bf \emph{1. Full program retrieval (Alg.~\ref{alg:cache}, line \ref{line:full}):}}
Given utterance $x$, construct an abstract utterance $\bar{x}$, retrieve the top $D$ abstract programs $\sA$ from the cache, compute the de-abstracted programs $Z$ using alignments from program tokens to utterance tokens, and add the $D$ programs to the {\em final} beam.
\newline
{\bf \emph{2. Program prefix retrieval (Alg.~\ref{alg:cache}, line \ref{line:prefix})}:} Here, we additionally consider prefixes of abstract programs to the beam, to further guide the search process. At each step $t$, let $B_t$ be the beam of decoded programs at step $t$.
For every abstract program $\bar{z} \in {\mathcal A}$ add the de-abstracted prefix $z_{1:t}$ to $B_t$ and expand $B_{t+1}$ accordingly. This allows the parser to potentially construct new programs that are not in the cache already.
This approach combats both spuriousness and the search challenge, because we add promising program prefixes to the beam that might have fallen off of it earlier. \reffig{cache} visualizes the caching mechanism.



A high-level overview of our entire approach for utilizing abstract examples at training time for both data augmentation and model training is given in \reffig{overview}.

\FigStar{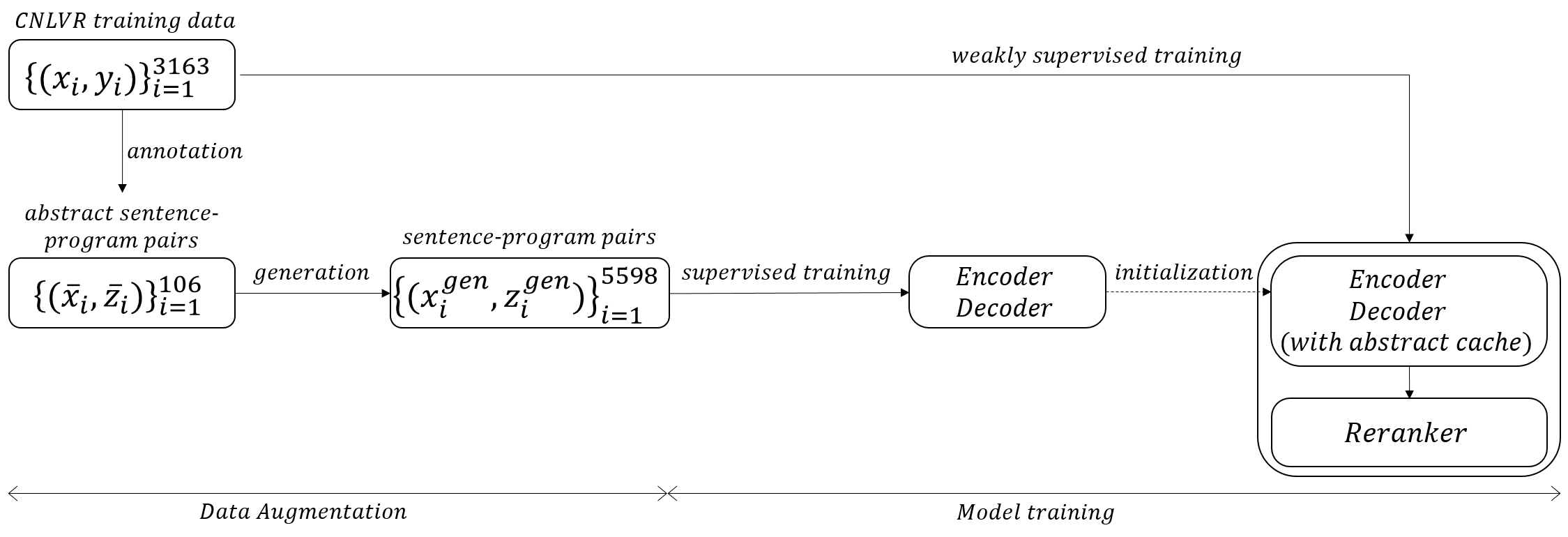}{0.4}{overview}{An overview of our approach for utilizing abstract examples for data augmentation and model training.}

\section{Experimental Evaluation} \label{sec:experiments}
\paragraph{Model and Training Parameters}
The Bi-LSTM state dimension is $30$. The decoder has one hidden layer of dimension
$50$, that takes the last 4 decoded tokens as input as well as encoder states.
Token embeddings are of dimension 12, beam size is $40$ and $D=10$ programs are
used in Algorithm \ref{alg:cache}. Word embeddings are initialized from CBOW
\cite{mikolov2013efficient} trained on the training data, and are then optimized
end-to-end. In the weakly-supervised parser we encourage exploration with meritocratic gradient updates  with $\beta=0.5$
\cite{guu2017bridging}.
In the weakly-supervised parser we warm-start the parameters with the supervised parser, as mentioned above.
For optimization, Adam is used  \cite{kingma2014adam}), with  learning rate of $0.001$, and mini-batch size of $8$. 

\paragraph{Pre-processing}
Because the number of utterances is relatively small for training a neural model, we take the following steps to reduce sparsity. We lowercase all utterance tokens, and also use their lemmatized form. We also use spelling correction to replace words that contain typos. After pre-processing we replace every word that occurs less than 5 times with an \texttt{UNK} symbol.

\paragraph{Evaluation}
We evaluate on the public development and test sets of \textsc{CNLVR} as well as on the hidden test set. The standard evaluation metric is accuracy, i.e., how many examples are correctly classified. In addition, we report \emph{consistency}, which is the proportion of utterances for which the decoded program has the correct denotation for all 4 images/KBs. 
It captures whether a model consistently produces a correct answer.

\paragraph{Baselines}
We compare our models to the \textsc{Majority} baseline that  picks the majority class (\textsc{True} in our case). We also compare to the state-of-the-art model reported by \newcite{suhr2017nlvr} when taking the KB as input, which is a maximum entropy classifier (\textsc{MaxEnt}). 
For our models, we evaluate the following variants of our approach:
\begin{itemize}[nosep,leftmargin=0.4cm]
\item \textsc{Rule}: The rule-based parser from \refsec{abs}. 
\item \textsc{Sup.}: The supervised semantic parser trained on augmented data as
  in \refsec{augmented} ($5,598$ examples for training and $560$ for validation).
\item \textsc{WeakSup.}: Our full  weakly-supervised semantic parser that uses abstract examples.
\item \textsc{+Disc}: We add a discriminative re-ranker (\refsec{model}) for both \textsc{Sup.} and \textsc{WeakSup.}
\end{itemize}

\paragraph{Main results}

\begin{table}[t]
\begin{center}
\scriptsize{
\begin{tabular}{l|c|c|c|c|c|c}
  & \multicolumn{2}{c|}{\textbf{Dev.}} & \multicolumn{2}{c|}{\textbf{Test-P}} & \multicolumn{2}{c}{\textbf{Test-H}}\\

 \textbf{Model} & \textbf{Acc.} & \textbf{Con.} & \textbf{Acc.} & \textbf{Con.} & \textbf{Acc.} & \textbf{Con.}\\
 \hline
 \textsc{Majority} & 55.3& - & 56.2 & -& 55.4 & - \\
 \textsc{MaxEnt} & 68.0 & - & 67.7 & - & 67.8 & - \\
 \hline
 \textsc{Rule} & 66.0 & 29.2 & 66.3 & 32.7 & - & -\\
 \textsc{Sup.} & 67.7 & 36.7 & 66.9 &38.3 & -& -\\
 \scriptsize{\textsc{Sup.+Disc}} & 77.7 & 52.4 & 76.6 &51.8	 & -& -\\
 \textsc{WeakSup.} & 84.3 & 66.3 & 81.7 & 60.1 & - & -\\
 \scriptsize{\textsc{W.+Disc}} & \textbf{85.7} & \textbf{67.4} & \textbf{84.0} & \textbf{65.0} & \textbf{82.5} & \textbf{63.9}\\
\end{tabular}}
\end{center}
\caption{Results on the development, public test (Test-P) and hidden test (Test-H) sets. For each model, we report both accuracy and consistency.}
\label{tab:main_res}
\end{table}

Table \ref{tab:main_res} describes our main results. Our weakly-supervised semantic parser with re-ranking (\textsc{W.+Disc}) obtains $84.0$ accuracy and $65.0$ consistency on the public test set and $82.5$ accuracy and $63.9$ on the hidden one, improving accuracy by $14.7$ points compared to state-of-the-art.
The accuracy of the rule-based parser (\textsc{Rule}) is less than $2$ points
below \textsc{MaxEnt}, showing that a semantic parsing approach is very suitable for this task. The supervised parser obtains better performance
(especially in consistency), and with re-ranking reaches $76.6$ accuracy,
showing that generalizing from generated examples is better than memorizing
manually-defined patterns. Our weakly-supervised parser significantly improves
over \textsc{Sup.}, reaching an accuracy of $81.7$ before
re-ranking, and $84.0$ after re-ranking (on the public test set). Consistency results show an even crisper
trend of improvement across the models. 



\subsection{Analysis}
\label{sec:analysis}

\begin{table}[t]
\begin{center}
\footnotesize{
\begin{tabular}{l|c|c}
  & \multicolumn{2}{c}{\textbf{Dev.}}\\
\textbf{Model} & \textbf{Acc.} & \textbf{Con.} \\
\hline

\textsc{Randomer} & 53.2 & 7.1\\ 
\textsc{$-$Abstraction} & 58.2 & 17.6\\ 
  \textsc{$-$DataAugmentation} & 71.4 & 41.2\\
 \textsc{$-$BeamCache}& 77.2 & 56.1\\  
 \textsc{$-$EveryStepBeamCache} & 82.3 & 62.2 \\ 
 \textsc{OneExampleReward} & 58.2 & 11.2\\
\end{tabular}}
\end{center}
\caption{Results of ablations of our main models on the development set. Explanation for the nature of the models is in the body of the paper.}
\label{tab:dev_res}
\end{table}

We analyze our results by running multiple ablations of our best model \textsc{W.+Disc} on the development set.

To examine the overall impact of our procedure, we trained a weakly-supervised parser from scratch without pre-training a supervised parser nor using a cache, which amounts to a re-implementation of the \textsc{Randomer} algorithm \cite{guu2017bridging}. We find that the algorithm is unable to bootstrap in this challenging setup and obtains very low performance. Next, we examined the importance of abstract examples, by pre-training only on examples that were manually annotated (utterances that match the $106$ abstract patterns), but with no data augmentation or use of a cache (\textsc{$-$Abstraction}). This results in  performance that is similar to the \textsc{Majority} baseline.

To further examine the importance of abstraction, we decoupled the two contributions, training once with a cache but without data augmentation for pre-training (\textsc{$-$DataAugmentation}), and again  with pre-training over the augmented data, but without the cache (\textsc{$-$BeamCache}). We found that the former improves by a few points over the \textsc{MaxEnt} baseline, and the latter performs comparably to the supervised parser, that is, we are still unable to improve learning by training from denotations.

Lastly, we use a beam cache without line \ref{line:prefix} in
Alg.~\ref{alg:cache} (\textsc{$-$EveryStepBeamCache}). This already results in
good performance, 
substantially higher than \textsc{Sup.} but is still
$3.4$ points worse than our best performing model on the development set. 

Orthogonally, to analyze the importance of tying the reward of all four examples that share an utterance, we trained a model without this tying, where the reward is 1 iff the denotation is correct (\textsc{OneExampleReward}). We find that spuriousness becomes a major issue and weakly-supervised learning fails.

\paragraph{Error Analysis}
We sampled 50 consistent and 50 inconsistent programs from the development set to analyze the weaknesses of our model. By and large, 
errors  correspond to utterances that are more complex syntactically and
semantically. In about half of the errors an object was described by two or more
modifying clauses: \emph{``there is a box with a yellow circle and three blue
items"}; or nesting occurred: \emph{``one of the gray boxes has exactly three
objects one of which is a circle"}. In these cases the model either ignored one
of the conditions, resulting in a program equivalent to \emph{``there is a box with three blue items"} for the first case, or applied composition operators wrongly, outputting an equivalent to \emph{``one of the gray boxes has exactly three circles"} for the second case. However, in some cases the parser succeeds on such examples and we found that 12\% of the sampled utterances that were parsed correctly had a similar complex structure. 
Other, less frequent reasons for failure were problems with cardinality interpretation, i.e. ,\emph{``there are 2"} parsed as \emph{``exactly 2"} instead of \emph{``at least 2"}; applying conditions to items rather than sets, e.g., \emph{``there are 2 boxes with a triangle closely touching a corner"} parsed as \emph{``there are 2 triangles closely touching a corner"}; and utterances with questionable phrasing, e.g.,  \emph{``there is a tower that has three the same blocks color"}.


Other insights are that the algorithm tended to give higher probability to the top ranked program when it is correct (average probability $0.18$), compared to cases when it is incorrect (average probability $0.08$), indicating that probabilities are correlated with confidence. In addition, 
sentence length is not predictive for whether the model will succeed: average sentence length of an utterance is $10.9$ when the model is correct, and $11.1$ when it errs.

We also note that the model was successful with sentences that deal with spatial relations, but struggled with sentences that refer to the size of shapes. This is due to the data distribution, which includes many examples of the former case and fewer examples of the latter.
\section{Related Work}
Training semantic parsers from denotations has been one of the most popular training schemes for scaling semantic parsers since the beginning of the decade. Early work focused on traditional log-linear models \cite{clarke10world,liang11dcs,kwiatkowski2013scaling}, but recently denotations have been used to train neural semantic parsers \cite{liang2017nsm,krishnamurthy2017neural,rabinovich2017abstract,cheng2017learning}.

Visual reasoning has attracted considerable attention, with datasets such as \textsc{VQA} \cite{antol2015vqa} and \textsc{CLEVR} \cite{johnson2017clevr}.
The advantage of \textsc{CNLVR} is that language utterances are both natural and compositional. Treating visual reasoning as an end-to-end semantic parsing problem has been previously done on \textsc{CLEVR} \cite{hu2017learning,johnson2017inferring}.


Our method for generating training data resembles data re-combination ideas in \citet{jia2016recombination}, where examples are generated automatically by replacing entities with their categories.

While spuriousness is central to semantic parsing when denotations are not very informative, there has been relatively little work on explicitly tackling it. \newcite{pasupat2015compositional} used manual rules to prune unlikely programs on the \textsc{WikiTableQuestions} dataset, and then later utilized crowdsourcing \cite{pasupat2016inferring} to eliminate spurious programs. \newcite{guu2017bridging} proposed \textsc{Randomer}, a method for increasing exploration and handling spuriousness by adding randomness to beam search and a proposing a ``meritocratic" weighting scheme for gradients. In our work we found that random exploration during beam search did not improve results while meritocratic updates slightly improved performance.

\section{Discussion}
In this work we presented the first semantic parser for the \textsc{CNLVR} dataset, taking structured representations as input. Our main insight is that in closed, well-typed domains we can generate abstract examples that can help combat the difficulties of training a parser from delayed supervision. First, we use abstract examples to semi-automatically generate utterance-program pairs that help warm-start our parameters, thereby reducing the difficult search challenge of finding correct programs with random parameters. Second, we focus on an abstract representation of examples, which allows us to tackle spuriousness and alleviate search, by sharing information about promising programs between different examples.
Our approach dramatically improves performance on \textsc{CNLVR}, establishing a new state-of-the-art.


In this paper, we used a manually-built high-precision lexicon to construct abstract examples. This is suitable for well-typed domains, which are ubiquitous in the virtual assistant use case.
In future work we plan to extend this work and automatically learn such a lexicon.
This can reduce manual effort and scale to larger domains where 
there is substantial variability on the language side.

\section*{Acknowledgements}
This research was partially supported by The Israel Science Foundation grant 942/16, and by the Yandex Initiative for Machine Learning.

\bibliography{all}
\bibliographystyle{acl_natbib}
 \end{document}